\documentclass{article}

\usepackage[final]{corl_2018}

\usepackage{graphicx}
\usepackage{amsmath,amssymb}
\usepackage{color}
\usepackage{multirow}
\usepackage[caption=false]{subfig}
\usepackage{tabulary}

\newcommand{\eg}{\textit{e}.\textit{g}. }
\newcommand{\bd}[1]{\textbf{#1}}
\newcolumntype{x}[1]{>{\centering\arraybackslash}p{#1pt}}
\newlength\savewidth\newcommand\shline{\noalign{\global\savewidth\arrayrulewidth
  \global\arrayrulewidth 1pt}\hline\noalign{\global\arrayrulewidth\savewidth}}
\newcommand{\tablestyle}[2]{\setlength{\tabcolsep}{#1}\renewcommand{\arraystretch}{#2}\centering\footnotesize}

\title{HDNET: Exploiting HD Maps for 3D Object Detection}

\author{
  Bin Yang$^{1,2}$ \quad Ming Liang$^{1}$ \quad Raquel Urtasun$^{1,2}$\\
  $^{1}$Uber Advanced Technologies Group \quad $^{2}$University of Toronto\\
  \texttt{\{byang10, ming.liang, urtasun\}@uber.com} \\
}

\begin{document}
\maketitle

%!TEX root = top.tex
\begin{abstract}
In this paper we show that High-Definition (HD) maps provide strong priors that can boost the performance and robustness of modern 3D object detectors.
Towards this goal, we design a single stage detector that extracts geometric and semantic features from the HD maps.
As maps might not be available everywhere, we also propose a map prediction module that estimates the map on the fly from raw LiDAR data.
We conduct extensive experiments on KITTI \cite{kitti} as well as a large-scale 3D detection benchmark containing 1 million frames, and show that the proposed map-aware detector consistently outperforms the state-of-the-art in both mapped and un-mapped scenarios.
Importantly the whole framework runs at 20 frames per second.
\end{abstract}

%!TEX root = top.tex
\section{Introduction}

Autonomous vehicles have the potential of providing cheaper and safer transportation. A typical autonomous system is composed of the following functional modules: perception, prediction, planning and control \cite{levinson2011towards}. Perception is concerned with detecting the objects of interest (\eg vehicles) in the scene and track them over time. The prediction module estimates the intentions and trajectories of all actors into the future. Motion planning is responsible for producing a trajectory that is safe, while control outputs the commands necessary for the self-driving vehicle to execute such trajectory.

3D object detection is a fundamental task in  perception systems.  
Modern 3D object detectors \cite{pixor, mv3d} exploit LiDAR as input as it provides good geometric cues and eases 3D localization when compared to camera-only approaches.
In the context of  real-time applications, single-shot detectors \cite{focal, ssd, yolo} have been shown to be more promising  than  proposal-based  methods \cite{faster, mv3d} as they are very efficient and can produce very accurate estimates.
However, object detection is far from solved as many challenges remain, such as dealing with occlusion and the sparsity of the LiDAR at long range.

Most self-driving systems have access to High-Definition (HD) maps that contain geometric and semantic information about the environment. While HD maps are widely used by motion planning systems \cite{seif2016, chen2016motion}, they are vastly ignored by perception systems \cite{holistic}.
In this paper we argue that HD maps provide strong priors that can boost the performance and robustness of modern object detectors. 
Towards this goal, we derive an efficient and effective  single-stage detector that operates in Bird's Eye View (BEV) and fuses LiDAR information with rasterized maps. Bird's eye view is a good representation for 3D LiDAR as it is amenable to efficient inference and  retains the metric space. 
Since HD maps might not be available everywhere, we also propose a map prediction module that estimates the map geometry and semantics from a single online LiDAR sweep.

Our experiments on the public KITTI BEV object detection benchmark \cite{kitti} and a large-scale 3D object detection benchmark TOR4D \cite{pixor, faf} show that we can achieve significant Average Precision (AP) gain on top of a state-of-the-art detector by exploiting HD maps. On TOR4D when HD maps are available, we achieve 2.42\%, 3.43\% and 5.49\% AP gains for ranges over 0-70 m, 30-50 m and 50-70 m respectively.  
On KITTI, where HD maps are unavailable, we show that when using  a pre-trained map prediction module (trained on a different continent) we can still get 2.87\% AP gain, surpassing all competing methods including those which also exploit cameras. Importantly, the proposed map-aware detector runs at 20 frames per second.
%!TEX root = top.tex
\section{Related Work}

%==========================================
\begin{figure*}[t]
\begin{center}
 \includegraphics[width=1.0\linewidth]{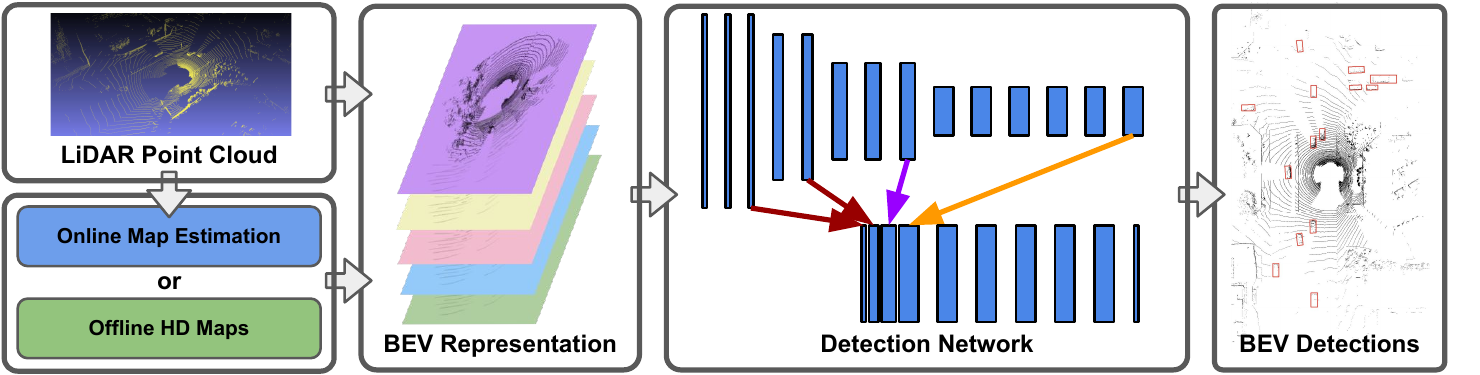}
\end{center}
   \caption{The overall architecture of the proposed map-aware single-stage 3D detector that operates on bird's eye view LiDAR representation.}
\label{fig:framework}
\end{figure*}
%===========================================

In this paper we show how modern 3D object detectors can benefit from HD maps. 
Here we revisit literatures on 3D object detection from point clouds, as well as works that exploit priors from maps.

\subsection{3D Object Detection in Point Clouds}
Some detectors \cite{vote3d, vote3deep, 3dfcn} search objects in 3D space densely via sliding window. Due to the sparsity of LiDAR point cloud, these approaches suffer from expensive computation, as many proposals are evaluated where there is no evidence. A more efficient representation of point clouds is to exploit 2D projections. MV3D \cite{mv3d} uses a combination of range view and bird's eye view projections to fuse multi-view information. PIXOR \cite{pixor} and FAF \cite{faf}  show superior performance in both speed and accuracy by exploiting the bird's eye view representation alone. Apart from grid-like representations, some works \cite{voxelnet, fpointnet} learn feature representations from un-ordered point sets for object detection. In addition to point cloud based detection, many works \cite{mv3d, fpointnet, avod} try to fuse data from multiple sensors to improve performance.

\subsection{Exploiting Priors from Maps}
Maps contain geographic, geometric and semantic priors that are useful for many tasks. \cite{holistic} leverages dense priors from large-scale crowd-sourced maps to build a holistic model that does joint 3D object detection, semantic segmentation and depth reconstruction. However, the priors are extracted by rendering a 3D world from the map, which is very time-consuming. In \cite{mattyus2015enhancing, mattyus2016hd}, the crowd-sourced maps are also used for fine-grained road segmentation.  3D object detectors often use the ground prior \cite{3doppami, birdnet}. However, they treat ground as a plane which is often inaccurate for curved roads. In this paper, we exploit the point-wise geometric ground prior as well as the semantic road prior for 3D object detection.
In \cite{zhang2013understanding, geiger20143d}, online estimated maps are used to reason the location of objects in 3D. However, the proposed generative model is very slow and thus not amenable for robotics applications. \cite{djuric2018motion} uses HD maps to predict the intention of different traffic actors like vehicles, pedestrians and bicyclists, which belongs to the more high-level prediction system.
%!TEX root = top.tex

\section{Exploiting HD Maps for 3D Object Detection}

HD maps are typically employed for motion planning but they are vastly ignored for perception systems.  
In this paper we bridge the gap and study how HD maps can be used as priors for modern 3D object detectors. Towards this goal, we derive a single-stage detector that can exploit both semantic and geometric information extracted from the maps (whether built offline or estimated online).
We refer the reader to Figure \ref{fig:framework} for the overall architecture of our proposed map-aware detector. 

%==========================================
\begin{figure*}[t]
\begin{center}
 \includegraphics[width=1.0\linewidth]{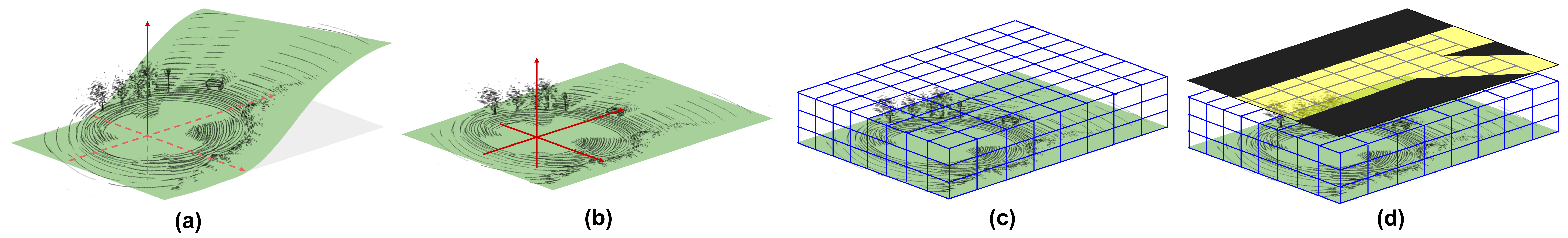}
\end{center}
   \caption{BEV LiDAR representation that exploits geometric and semantic HD map information. (a) The raw LiDAR point cloud. (b) Incorporating geometric ground prior. (c) Discretization of the LiDAR point cloud. (d) Incorporating semantic road prior.}
\label{fig:map}
\end{figure*}
%===========================================

\subsection{Input Representation}
We project the LiDAR data to bird's eye view (BEV) as this provides a compact representation that enables efficient inference. Note that this is a good representation for our application as vehicles drive on the ground. Figure \ref{fig:map} illustrates how we incorporate priors from HD maps into our BEV LiDAR representation.

We treat the $Z$ axis as feature channel and exploit 2D convolutions. This is beneficial as they are much more efficient than 3D convolutions. 
It is therefore important to have  discriminant features along the $Z$ axis. 
However, LiDAR point clouds often suffer from translation variance along the $Z$ axis due to the slope of the road, particularly at range. For example, 1 degree of road slope would lead to 1.22 meters offset in $Z$ axis at 70 meters distance. To make things worse, standard LiDAR sensors have very sparse returns after 50 meters.

To remedy this translation variance, we exploit  detailed ground information from the HD maps. Specifically, given a LiDAR point cloud $\{(x_i, y_i, z_i)\}$, we query the ground point at each location $(x_i, y_i)$ from the map, denoted as $(x_i, y_i, z^0_i)$ and replace the absolute distance $z_i$ with  the distance relative to the ground $z_i - z^0_i$. 
We then discretize the resulting representation into a 3D occupancy  grid enhanced to also contain  LiDAR intensity features following \cite{pixor}. Specifically, we first define the 3D dimension $L \times W \times H$ of the scene that we are interested in. 
We then compute binary occupancy  maps at a resolution of $d_{L} \times d_{W} \times d_{H}$, and compute the intensity feature map at a resolution of $d_{L} \times d_{W} \times H$.
The discretized LiDAR BEV representation is of  size  $\frac{L}{d_L} \times \frac{W}{d_W}$, with $\frac{H}{d_H}+3$ channels\footnote{Two additional occupancy channels are added to cover points outside the height range.}.

The LiDAR provides  a full scan of the surrounding environment, containing moving objects and static background (\eg roads and buildings). However, in the context of self driving cars we mainly care about moving objects on the road. Motivated by this, we exploit the semantic road mask available on  HD maps as prior knowledge about the scene.
Specifically, we extract the road layout information from the HD maps and rasterize it onto the bird's eye view as a binary  channel at the same resolution as the discretized LiDAR representation. We then concatenate the road mask together with the LiDAR representation along the channels to create our input representation.

\subsection{Network Structure}
\begin{figure*}[t]
\begin{center}
\includegraphics[width=0.42\linewidth]{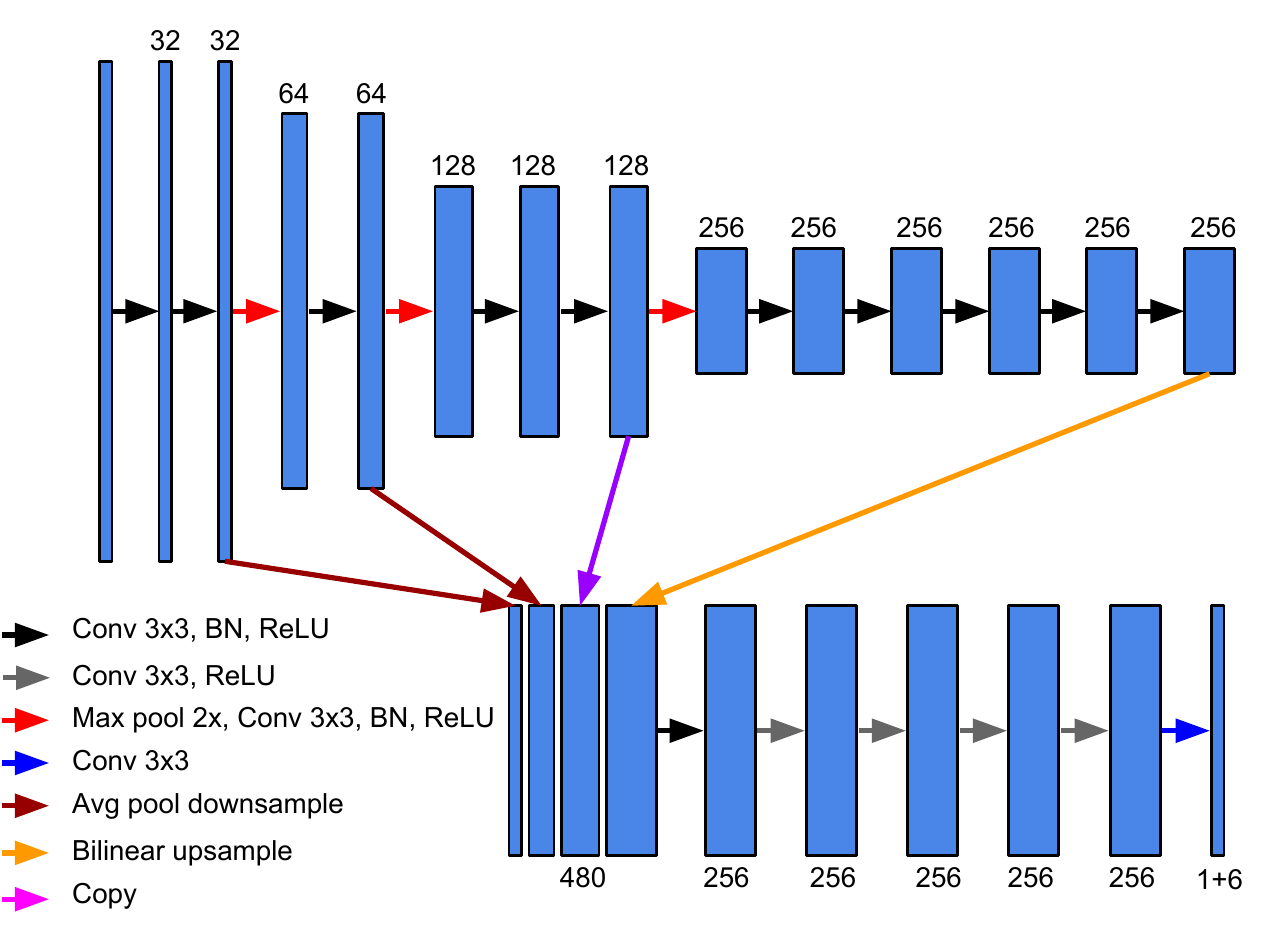} \quad\quad \includegraphics[width=0.52\linewidth]{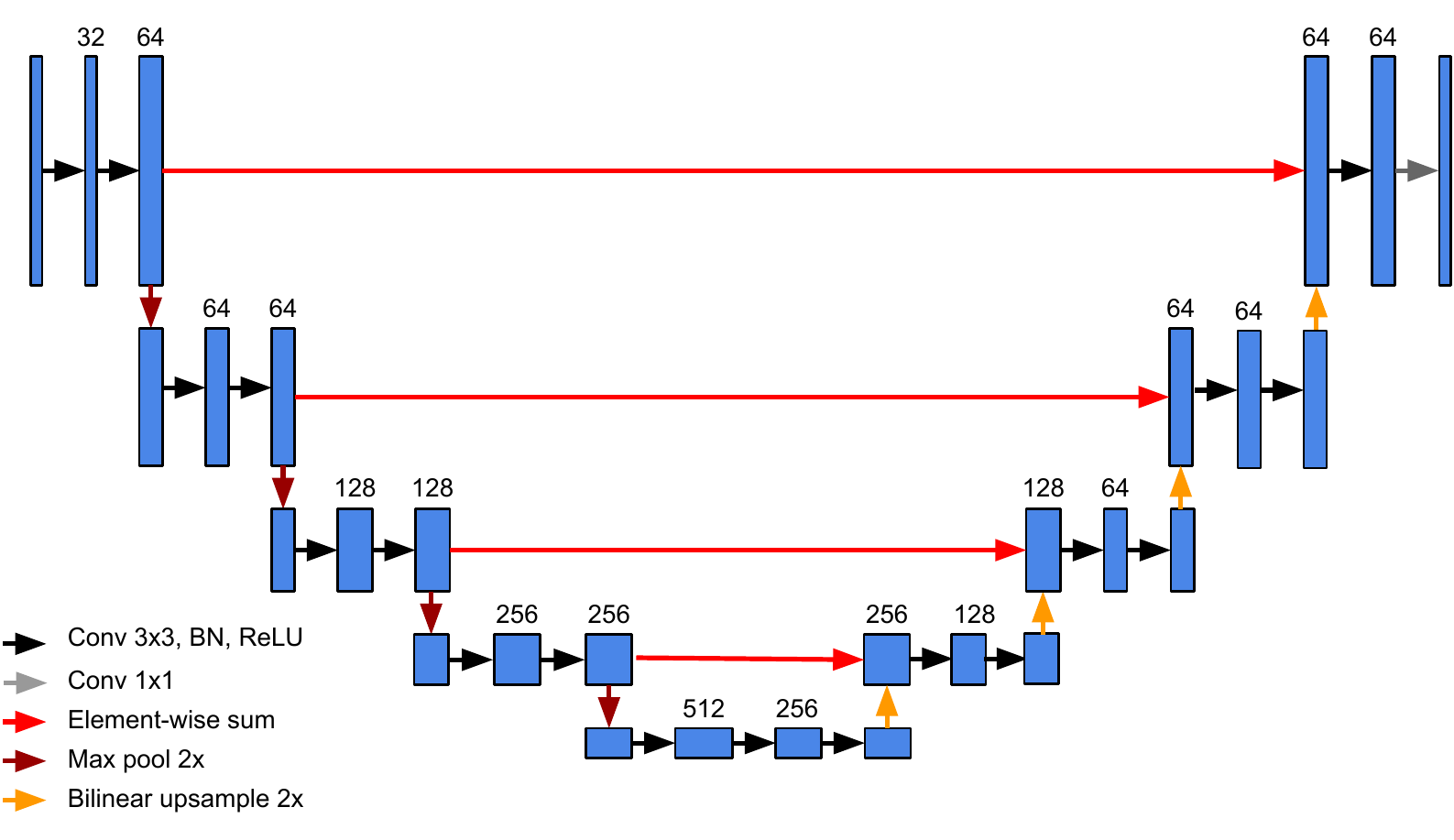}
\end{center}
   \caption{Network structures for object detection (left) and online map estimation (right).}
\label{fig:network}
\end{figure*}

We adopt a fully convolutional network for single-stage dense object detection. Figure \ref{fig:network} (left) shows the detection network, which is composed of two parts: a \emph{backbone network} that extracts multi-scale features and a \emph{header network} that outputs pixel-wise dense detection estimates.

\bd{Backbone network:} consists of four convolutional blocks, each having \{2, 2, 3, 6\} \texttt{conv2D} layers with filter number \{32, 64, 128, 256\}, filter size 3, and stride 1. We apply batch normalization \cite{bn} and ReLU  \cite{relu} after each \texttt{conv2D} layer. After each of the first three convolutional blocks there's a \texttt{MaxPool} layer with filter size 3 and stride 2. Multi-scale features are generated by resizing and concatenating feature maps from different blocks. The total downsampling rate of the network is 4.

\bd{Header network:} consists of five \texttt{conv2D} layers with filter number 256, filter size 3 and stride 1, followed by the last \texttt{conv2D} layer that outputs pixel-wise dense detection estimates $(p, cos(2\theta), sin(2\theta), dx, dy, \log w, \log l)$. $p$ denotes the confidence score for object classification, where the rest denotes the geometry of the detection. Specifically, we parameterize the object orientation (in $XY$ plane) by $(cos(2\theta), sin(2\theta))$ with a period of $\pi$ so that we do not distinguish between heading forward and backward\footnote{Because they are the same in terms of IoU computation.}. $(dx, dy, \log w, \log l)$ are commonly used parameterization \cite{frcn} for object center offset and size. The advantages of producing dense detections are two-fold: it is efficient to compute via convolutions and  the maximum recall rate of objects is 100\%.

\subsection{Learning and Inference}
Adding priors (semantic prior in particular) at the input level has both advantages as well as challenges. On one hand, the network can fully exploit the interactions between priors and raw data; on the other hand, this may lead to overfitting issues, and potentially poor results when HD maps are unavailable or noisy. In practice, having a detector that works regardless of map availability is important. Towards this goal, we apply {\it data dropout} on the semantic prior, which randomly feeds an empty road mask to the network during training. Our experiments show 	 that data dropout largely improves the model's robustness to map availability. 

We employ the commonly used multi-task loss \cite{frcn} to train the detection network. Specifically, we use focal loss \cite{focal} on the classification task and smooth $\ell_1$ loss on the regression task. The total loss is simply the combination of classification loss summed over all pixels and regression loss summed over all positive pixels (no sampling strategy is required). We determine positive and negative samples according to pixel distance to the nearest ground-truth box center. We also normalize the regression targets to have zero mean and unit variance before training.

During inference, we pick all pixels on the output feature map whose confidence score is above a certain threshold, and decode them into oriented bounding boxes. Non-Maximum-Suppression (NMS) with 0.1 Intersection-Over-Union (IoU) threshold is then applied to get the final detections.

%==========================================
%===========================================

\section{Online Estimation of HD Maps}
So far we have shown how  to exploit geometric and semantic priors from the already built HD maps. However, such maps might not be available everywhere. 
To tackle this problem, we create  the map priors  online from a single LiDAR sweep. In our scenario, there is no need to estimate  dense 3D HD maps. Instead, we only have to predict the BEV representation of the geometric and semantic priors (shown in Figure \ref{fig:map}). In this way, the estimated map features can be seamlessly integrated into the current framework.

\bd{Network structure:} We estimate online maps from LiDAR point clouds via two separate neural networks that tackle ground estimation and road segmentation respectively. The network structure for these two tasks are the same U-Net structure \cite{unet} as shown in Figure \ref{fig:network} (right). This  has the advantages of retaining low-level details and producing high-resolution predictions.

\bd{Ground estimation:} Existing approaches  \cite{birdnet, 3doppami}  assume that the ground is a plane. This is, however,   inaccurate, particularly at  long range. Instead,  in this paper we predict the ground height value for every location on the BEV space. This results in a much more precise estimate.
During training, we extract ground-truth ground from existing HD maps, and compute $\ell_2$ loss only on locations where there exist LiDAR points. Empirically we find that ignoring empty locations during training leads to better performance.

\bd{Road segmentation:} We predict pixel-wise BEV road segmentation as the estimated road prior. During training we use the rasterized road mask as ground-truth label and train the network with cross-entropy loss summed over all locations. We refer the reader to Figure \ref{fig:map} for an illustration.

%!TEX root = top.tex
%===============================
\begin{table*}[t]
\begin{center}
%\begin{small}
\begin{tabular}{c|c|cc|c|ccc}
\multirow{2}{*}{Method} & \multirow{2}{*}{HD Maps} & \multirow{2}{*}{Ground} & \multirow{2}{*}{Road} & \multirow{2}{*}{Time (ms)} & \multicolumn{3}{c}{AP@0.7IoU (\%)}\\
\cline{6-8}
& & & & & 0-70 m & 30-50 m & 50-70 m\\
\shline
FAF \cite{faf} & no & N/A & N/A & 30 & 77.48 & N/A & N/A \\
PIXOR \cite{pixor}& no & N/A & N/A & 17 & 78.82 & N/A & N/A \\
PIXOR++ & no & N/A & N/A & 17 & 81.78 & 78.34 & 61.04 \\
\hline
\multirow{3}{*}{HDNET (online)} & \multirow{3}{*}{no}  & $\surd$ &  & 32 & +0.91 & +1.69 & +0.92 \\
 & &  & $\surd$ & 32 & +0.88 & +1.54 & +1.87 \\
 & & $\surd$ & $\surd$ & 47 & \bd{+1.33} & \bd{+2.34} & \bd{+1.88} \\
 \hline
 \multirow{3}{*}{HDNET (offline)} & \multirow{3}{*}{yes} & $\surd$ &  & 17 & +1.58 & +2.29 & +3.24 \\
 & &  & $\surd$ & 17 & +1.05 & +1.53 & +2.52 \\
 & & $\surd$ & $\surd$ & 17 & \bd{+2.42} & \bd{+3.43} & \bd{+5.49}\\
\end{tabular}
%\end{small}
\end{center}
\caption{Evaluation results on TOR4D benchmark at different ranges. \emph{PIXOR++} is the baseline detector without exploiting HD maps. \emph{HDNET (online)} exploits online estimated map priors on top of the baseline, while \emph{HDNET (offline)} extracts priors from offline built HD maps.}
\label{tab:results}
\end{table*}
%================================

%==========================================
\begin{figure*}[t]
\begin{center}
%\fbox{\rule{0pt}{1.5in} \rule{0.95\linewidth}{0pt}}
 \includegraphics[width=1.0\linewidth]{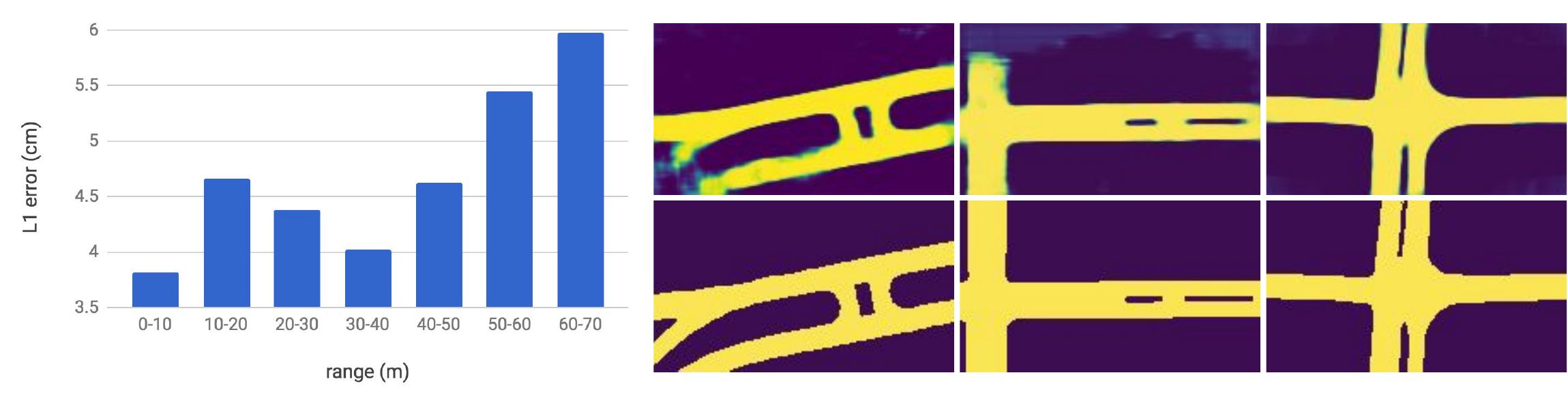}
\end{center}
   \caption{Evaluation of online map estimation. Left: range-wise error of ground height estimation. Right: the predicted road mask (1st row) versus the ground-truth road mask (2nd row).}
\label{fig:online}
\end{figure*}
%===========================================

\section{Experiments}
We validate the proposed method on two benchmarks: a large-scale 3D object detection benchmark TOR4D \cite{pixor, faf}, which has over one million frames as well as corresponding HD maps; and the public KITTI BEV object detection benchmark \cite{kitti}. We denote our baseline detector without exploiting map priors as PIXOR++ since it follows the overall architecture of PIXOR \cite{pixor}, and denote the map-aware detector which exploits map priors as HDNET.

\subsection{Evaluation on TOR4D Benchmark}
\bd{Implementation details:} We discretize the LiDAR point cloud within the following 3D region $x\in[-70.4, 70.4]$, $y\in[-40, 40]$, $z\in[-2, 3.4]$ in vehicle coordinate system (with the ego-car located at the origin). We use a discretization resolution of 0.2 meter in all three axes. The detection network is trained with stochastic gradient descent with momentum for 1.5 epochs with a batch size of 32. The initial learning rate is 0.02 and is multiplied by 0.1 after 1 and 1.4 epochs.

\bd{Baseline PIXOR++ detector:} The baseline detector achieves 81.78\% AP within 0-70 meters, outperforming FAF \cite{faf} and PIXOR \cite{pixor} by 4.3\% and 2.96\% respectively. What's more, the baseline detector runs at 17 ms per frame. In terms of performance at different ranges, we observe an AP drop of 3.44\% and 20.74\% when evaluated at 30-50 m range and 50-70 m range. This suggests that long range detection is significantly difficult.

\bd{Online map estimation:} Figure \ref{fig:online} shows the evaluation results of the online map prediction module. In ground height estimation task, we are able to achieve $<5$ cm L1 error pixel-wise within 50 m range. When the range becomes longer, the error increases since we have more sparse LiDAR observations there. In road segmentation task, we achieve 97.70\% pixel-wise accuracy and 92.86\% IoU on the validation set. Note that these results are from one single LiDAR sweep only.

\bd{Map-aware HDNET detector:} As seen from Table \ref{tab:results}, HDNET outperforms the baseline PIXOR++ by 1.33\%/2.42\% in settings with online/offline maps. When it comes to longer range, the gain in both settings increases, especially for the offline one (with 5.49\% gain in 50-70 m). This shows that map priors are especially helpful for the more difficult long range detection. Comparing geometric and semantic priors in both settings, we have three main observations: (1) geometric prior generally helps more than semantic prior, while in online setting the advantage diminishes (especially in long range) due to larger errors in online ground height estimation; (2) geometric prior and semantic prior are very complementary to each other; (3) in all settings and all ranges, we show that exploiting map priors could bring consistent gain over a state-of-the-art detector.

%===============================
\begin{table*}[t]
\begin{center}
%\begin{small}
\addtolength{\tabcolsep}{-0pt}
\begin{tabular}{c|c|c|ccc}
\multirow{2}{*}{Method} & \multirow{2}{*}{Input} & \multirow{2}{*}{Time (ms)} & \multicolumn{3}{c}{AP@0.7IoU (\%)}\\
\cline{4-6}
 & & & easy & moderate & hard \\
\shline
BirdNet~\cite{birdnet} & LiDAR & 110 & 75.52 & 50.81& 50.00 \\
MV3D~\cite{mv3d} & LiDAR & 240 & 85.82 & 77.00 & 68.94 \\
PIXOR~\cite{pixor} & LiDAR & \bd{35} & 81.70 & 77.05 & 72.95 \\
VoxelNet~\cite{voxelnet} & LiDAR & 225 & 89.35 & 79.26 & 77.39 \\
%NVLidarNet & LIDAR & 0.1 & n/a & n/a & n/a & 84.44 & 80.04 & 74.31 \\
%F-PC\_CNN~\cite{fpccnn} & LIDAR+Img & 0.5 & 60.06 & 48.07 & 45.22 & 83.77 & 75.26 & 70.17 \\
MV3D~\cite{mv3d} & LiDAR + Image & 360 & 86.02 & 76.90 & 68.49 \\
%AVOD-FPN~\cite{avod} & LIDAR+Img & ImageNet & 0.1 & 88.53 & 83.79 & 77.90\\
F-PointNet~\cite{fpointnet} & LiDAR + Image & 170 & 88.70 & 84.00 & 75.33 \\
AVOD~\cite{avod} & LiDAR + Image & 80 & 86.80 & 85.44 & 77.73 \\
ContFuse~\cite{contfuse} & LiDAR + Image & 60 & 88.81 & 85.83 & 77.33 \\
\hline
PIXOR++ & LiDAR & \bd{35} & \bd{89.38} & 83.70 & 77.97 \\
HDNET & LiDAR + Map & 50 & 89.14 & \bd{86.57} & \bd{78.32} \\
\end{tabular}
%\end{small}
\end{center}
\caption{Evaluation results on KITTI BEV object detection benchmark (car). \emph{PIXOR++} is the baseline detector, while \emph{HDNET} exploits map priors from an online map estimation module pre-trained on TOR4D dataset.}
\label{tab:kitti}
\end{table*}
%================================
\subsection{Evaluation on KITTI BEV Benchmark}
\bd{Implementation details:} We set the region of interest for the point cloud to $[0, 70.4] \times [-40, 40] \times [-3, 1]$ meter and use a discretization resolution of $0.1 \times 0.1 \times 0.2$ meter. Following \cite{pixor}, we use the same ResNet \cite{resnet2} based network structure that's more robust to overfitting. Since KITTI has only thousands of training frames and we do not use pre-trained weights for the network, we apply the following data augmentation during training. For each frame of LIDAR point cloud, we apply random scaling ($0.9\sim1.1$ for all 3 axes), translation ($-5\sim5$ meter for $X$ and $Y$ axes) and rotation ($-5\sim5$ degrees along $Z$ axis). KITTI only annotates objects within the camera field of view (FOV), therefore we ignore all pixels outside the camera FOV during training. We use $\alpha=0.75$ and $\gamma=0.5$ in the focal loss. We train the network with stochastic gradient descent with momentum for 50 epochs with a batch size of 16 frames. The initial learning rate is 0.01 and we decay it by $0.1$ after 30 and 45 epochs respectively.

\bd{BEV object detection:} We compare PIXOR++ and HDNET with other state-of-the-art detectors on KITTI BEV object detection benchmark and show the evaluation results in Table \ref{tab:kitti}. Our baseline detector PIXOR++ has already surpassed all the competing LiDAR based detectors, outperforming the second best VoxelNet \cite{voxelnet} by 4.44\% AP in moderate setting while being the fastest in runtime (35 ms). Because KITTI doesn't have HD maps information, we transfer the map prediction module trained on TOR4D to KITTI without any fine-tuning. Note that we do not exploit any object labels from the TOR4D dataset. With map priors added, HDNET achieves an absolute 2.87\% AP gain in moderate setting, setting the new record on the benchmark. Note that HDNET even outperforms other competitors that exploit camera images (\eg ContFuse \cite{contfuse}) or additional object labels from outside datasets (\eg F-PointNet \cite{fpointnet}), which highlights the long-ignored value of HD maps in 3D object detection.

\subsection{Ablation Studies}
We conduct four ablation studies investigating different ways to exploit the geometry and semantic priors, the effect of data dropout on robustness to unavailable maps, and the detection performance with regard to different ranges using priors from online estimated maps or offline built maps.

\subsubsection{Ground prior}
We model the geometric prior as a point-wise ground surface, which is more flexible and accurate than commonly used ground plane assumption \cite{3doppami, birdnet}. We compare these two design choices. For the ground plane counterpart, we use the publicly available ground plane results from 3DOP \cite{3doppami}, which were generated by road segmentation and RANSAC fitting. In contrast, we directly predict the point-wise ground. The way how the ground prior is used in detection is the same for these two methods. We evaluate these two priors using HDNET on KITTI validation set (without data augmentation) and show the results in Table \ref{tab:ablation:ground}. Both methods can boost the detection performance of the baseline, while our point-wise ground parameterization achieves significantly more gain.

\subsubsection{Road prior}
We incorporate semantic road prior into the input representation of the detector, while there are methods that use such prior at the output level. Two variants are created for output level fusion: multi-task learning and output masking. For multi-task learning, we add another output branch to the detection network to predict the road mask during training. For output masking, we mask the detection output with the road mask during both training and testing stages. We show the comparative results using the road prior extracted from offline HD maps in Table \ref{tab:ablation:road}, where we see that our input fusion performs the best among all three methods. For multi-task learning, we observe limited performance gain from jointly learning road mask and object detection. Output masking hurts the detection performance as errors in road prior is unrecoverable.

%================================
\begin{table*}[t]\centering\vspace{-3mm}
% Sub-tab first row
\subfloat[Ground prior \label{tab:ablation:ground}]{
\tablestyle{2pt}{1.05}
\begin{tabular}{c|ccc}
Method & AP@easy & AP@mod. & AP@hard \\
\shline
baseline & 87.31 & 79.29 & 76.02 \\
+ground plane (3DOP \cite{3doppami}) & 86.67 & 79.23 & 79.17 \\
\bd{+ground surface (ours)} & \bd{88.45} & \bd{82.60} & \bd{80.43} \\
\end{tabular}}\hspace{10mm}
\subfloat[Road prior \label{tab:ablation:road}]{
\tablestyle{2pt}{1.05}
\begin{tabular}{c|c}
Method & AP \\
\shline
multi-task learning & +0.25 \\
output masking & -1.14 \\
\bd{input fusion (ours)} & \bd{+2.42} \\
\end{tabular}}

% Sub-tab second row
\subfloat[Data dropout on road prior  \label{tab:ablation:dropout}]{
\tablestyle{1.5pt}{1.05}
\begin{tabular}{c|ccc}
Map Avail. & 0\% & 50\% & 100\% \\
\shline
w/o dropout & -11.57 & -6.37 & +0.77 \\
\bd{w/ dropout} & \bd{-0.20} & \bd{+0.34} & \bd{+0.84} \\
\end{tabular}}\hspace{5mm}
\subfloat[Detection performance at different ranges \label{tab:ablation:range}]{
\tablestyle{1.5pt}{1.05}
\begin{tabular}{c|ccccccc|c}
Range & 0-10 & 10-20 & 20-30 & 30-40 & 40-50 & 50-60 & 60-70 & 0-70 \\
\shline
online map & -0.15 & +0.44 & +1.17 & +2.45 & +1.88 & +0.97 & +1.03 & +1.33 \\
offline map & -0.09 & +0.65 & +1.45 & +2.93 & +3.71 & +4.25 & +6.38 & \bd{+2.42}\\
\end{tabular}}

\caption{Ablation studies on different ways to exploit ground prior and road prior, robustness to unavailable maps, and range-wise detection performances. All numbers are AP (\%) values.}
\label{tab:ablation}
\end{table*}
%================================

\subsubsection{Data dropout}
Data dropout is applied to the map prior during training stage to improve the model's robustness to missing maps. To validate the effectiveness of data dropout, we train two models with 100\% map availability. The only difference is that one model is trained with data dropout applied to the road prior, while the other model doesn't use data dropout. During testing, we manually control the availability of road prior at \{0\%, 50\%, 100\%\} and compare the performance. 

The evaluation results are shown in Table \ref{tab:ablation:dropout}, where we see that without data dropout, AP drops by 11.57\%/6.37\% at 0\%/50\% map availability. However, when data dropout is introduced, the map-aware detector is able to surpass the baseline when map's available; and when map's unavailable, the performance is almost as good as the baseline.

\subsubsection{Performance at various ranges}
We define the range as the distance from the object to the ego-car on the $XY$ plane. We divide the full 70 meters range into 7 bins, and when evaluating on each range bin we ignore the detections and labels outside this range bin. We conduct the fine-grained evaluation in both online and offline settings and show results in Table \ref{tab:ablation:range}. 

For offline setting, we see that the gain increases as the object range becomes larger. At $>40$ meters range, the gain is larger than 3\%. This makes sense as the map priors help more when we have fewer LiDAR observations. For online setting, we have the similar trending in near range ($<40$ meters), but the gain starts to decrease at long ranges ($>40$ meters). The reason is that map priors are estimated from single LiDAR sweep, which is very sparse in long range.

%======================================
\begin{figure*}[t]
\begin{center}
\frame{\includegraphics[width=0.25\linewidth]{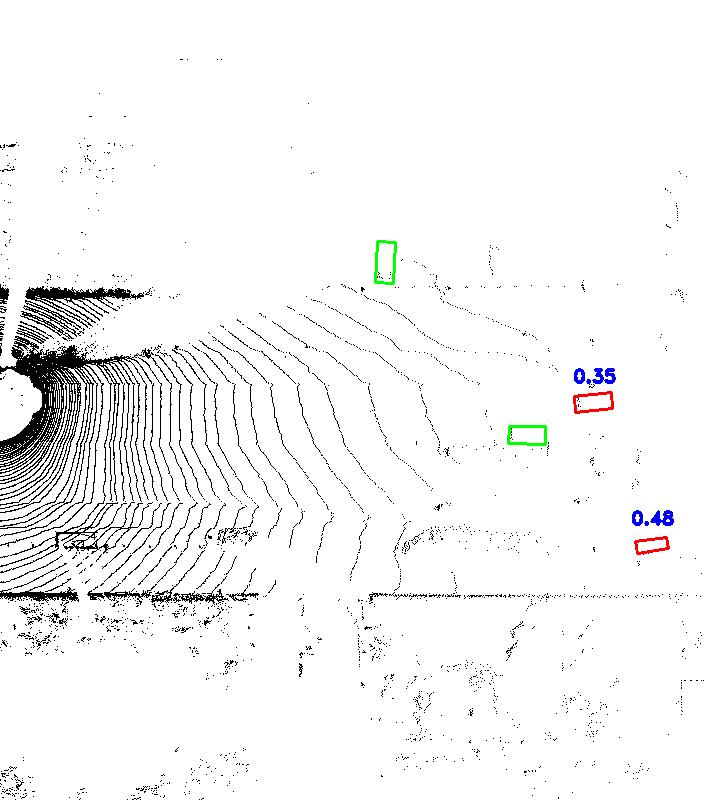}}\frame{\includegraphics[width=0.25\linewidth]{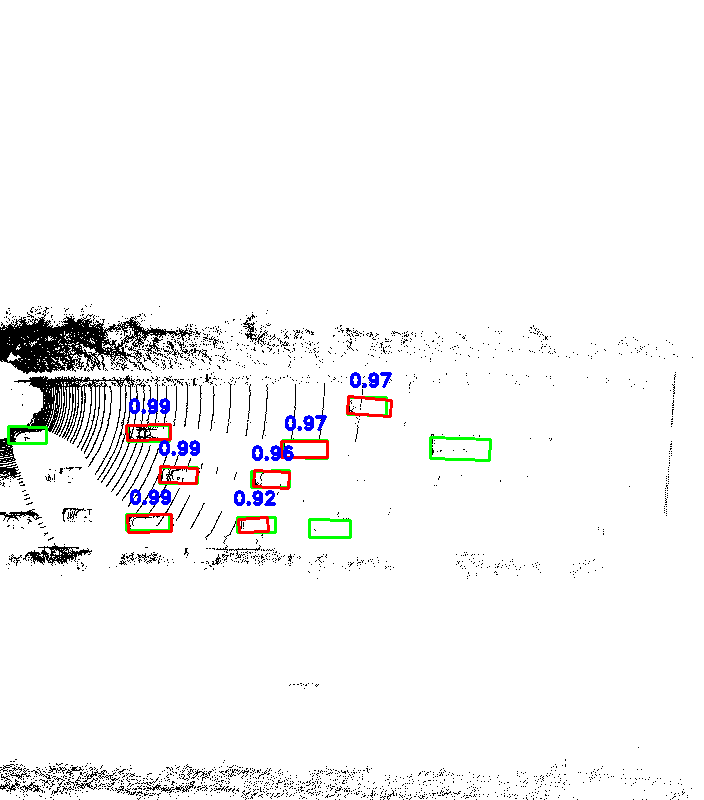}}\frame{\includegraphics[width=0.25\linewidth]{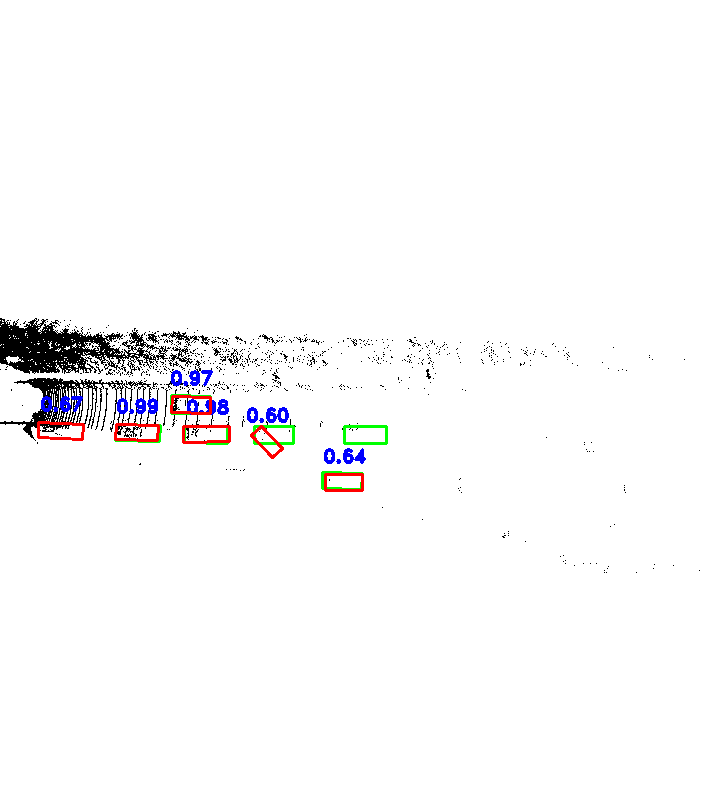}}\frame{\includegraphics[width=0.25\linewidth]{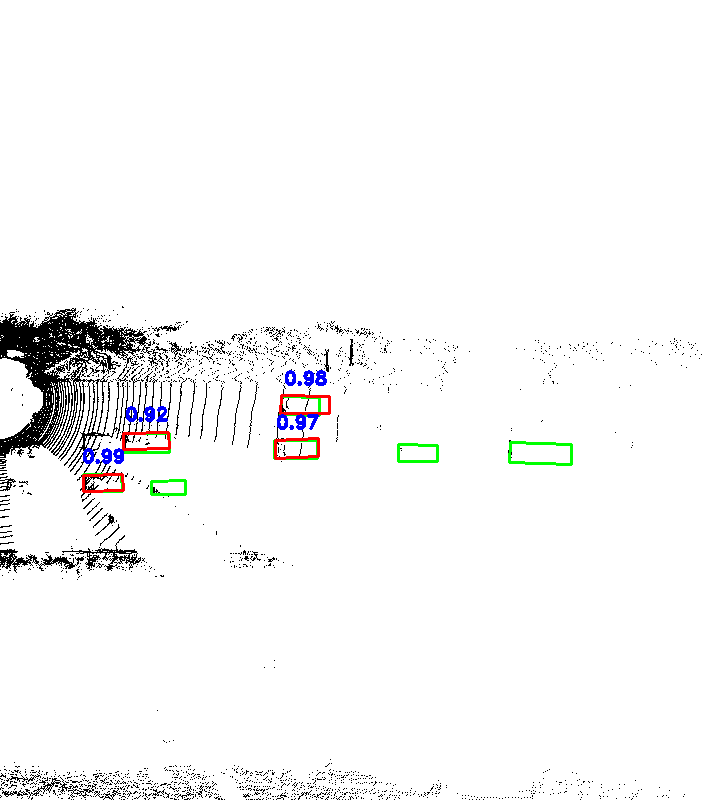}}\\
\frame{\includegraphics[width=0.25\linewidth]{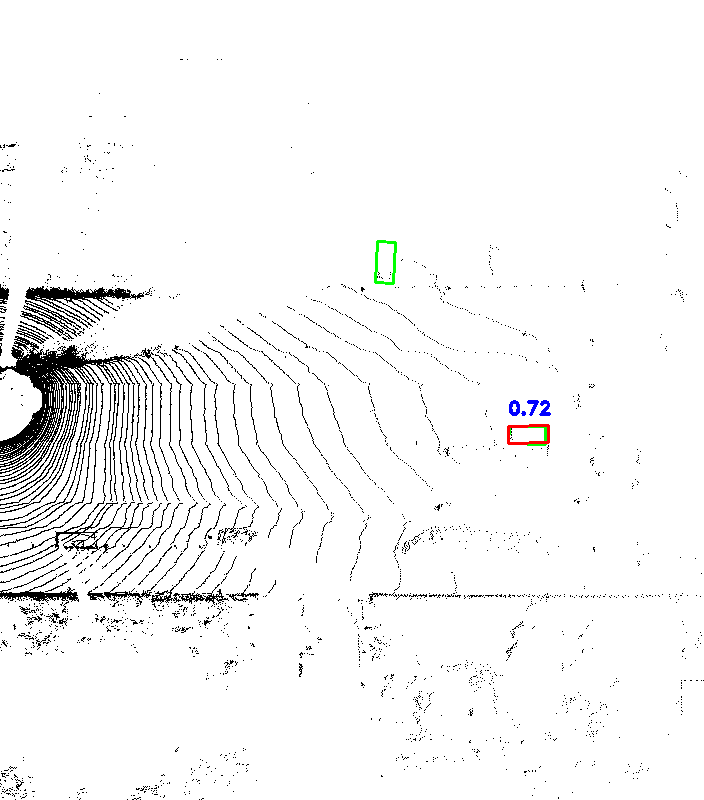}}\frame{\includegraphics[width=0.25\linewidth]{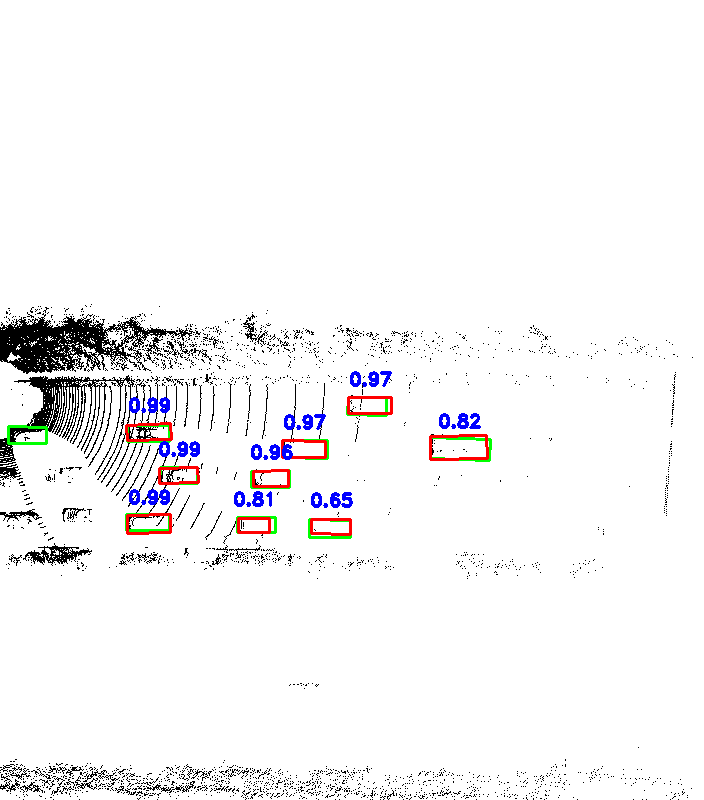}}\frame{\includegraphics[width=0.25\linewidth]{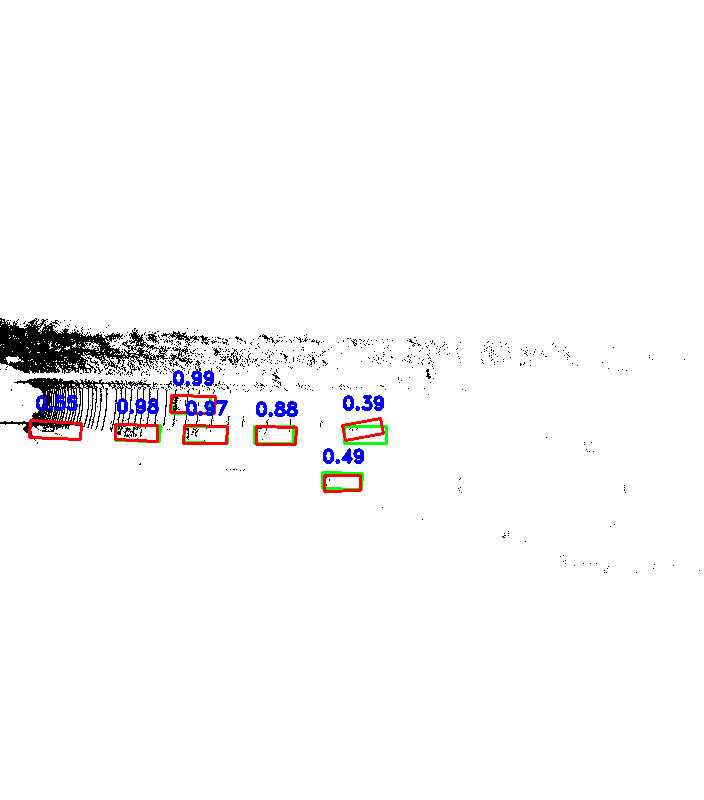}}\frame{\includegraphics[width=0.25\linewidth]{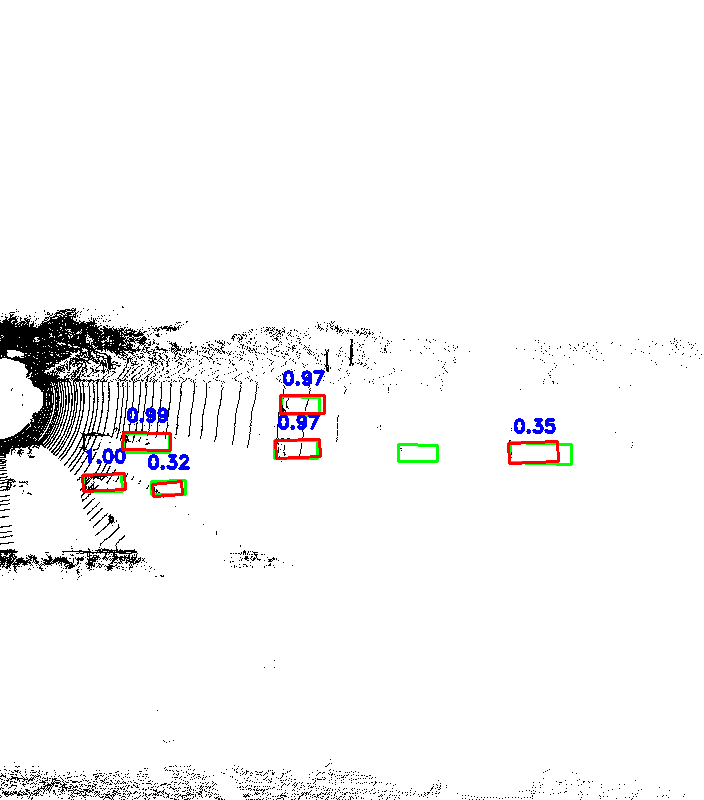}}
\end{center}
   \caption{Qualitative comparison of baseline PIXOR++ detector without map (1st row) and HDNET with online map (2nd row) on KITTI BEV object detection benchmark validation set. We show ground-truth labels in green, detection results in red, and detection scores in blue.}
\label{fig:demo}
\end{figure*}
%======================================

\subsection{Qualitative Results}
We compare some qualitative results from two detector variants: the baseline PIXOR++ detector without map, and the HDNET detector with online map estimation on the validation set of KITTI BEV object detection benchmark in Figure \ref{fig:demo}. 

From the figure we can easily identify false positives and false negatives in long range for PIXOR++. However, when it comes to HDNET, the detection accuracy and localization precision at long range increases remarkably. This shows that exploiting map priors does help the long range detection a lot.
%!TEX root = top.tex
\section{Conclusion}

In this paper we address the problem of how to exploit HD maps information to boost the performance of modern 3D object detection systems in the context of autonomous driving. We identify the geometric and semantic priors in HD maps, and incorporate them into the bird's eye view LiDAR representation. Because HD maps are not available everywhere, we also propose a map prediction module that estimates both map priors online from single LiDAR sweep. Experimental results on the public KITTI BEV object detection benchmark and a large-scale 3D object detection benchmark show that the proposed map-aware detector consistently outperforms the baseline detector whether the HD maps are available or not. The whole framework also runs at over 20 frames per second due to the use of BEV representation as well as the single-stage detection framework.

\clearpage
\bibliography{bib}

\end{document}